\documentclass[runningheads]{llncs}
\usepackage[utf8]{inputenc}
\usepackage[english]{babel}
\usepackage[T1]{fontenc}
\usepackage{graphicx}
\usepackage{amsmath}
\usepackage{amssymb}

\usepackage{booktabs}
\usepackage{caption}
\usepackage{float}
\usepackage{multirow}
\usepackage{subcaption}

\usepackage[dvipsnames]{xcolor}

\usepackage{hyperref}
\hypersetup{
    colorlinks=true,
    citecolor=blue,
    linkcolor=blue,
    urlcolor=blue
}
\renewcommand\UrlFont{\color{blue}\rmfamily}
\graphicspath{ {./paper_v2/images} }

\definecolor{pink}{RGB}{218, 3, 174}
\definecolor{mygreen}{RGB}{0, 128, 0}

\definecolor{grey}{RGB}{190, 173, 161}
\definecolor{greyL}{RGB}{179, 179, 179}
\definecolor{greyM}{RGB}{215, 215, 215}
\definecolor{greyLL}{RGB}{230, 230, 230}
\definecolor{redL}{RGB}{176, 0, 0}
\definecolor{redD}{RGB}{103, 10, 10}
\definecolor{bitter}{RGB}{253, 124, 110}
\definecolor{pink}{RGB}{255, 116, 253}
\definecolor{pinkL}{RGB}{255, 209, 220}
\definecolor{purpleL}{RGB}{126, 0, 89}
\definecolor{purpleLL}{RGB}{238, 190, 241}
\definecolor{purpleD}{RGB}{66, 0, 66}
\definecolor{magenta}{RGB}{148, 0, 211}
\definecolor{blueL}{RGB}{0, 140, 240}
\definecolor{blueLL}{RGB}{171, 205, 239}
\definecolor{malva}{RGB}{145, 95, 109}
\definecolor{greenIrland}{RGB}{76.0, 187, 23}
\definecolor{greenD}{RGB}{10.0, 69, 0}
\definecolor{maple}{RGB}{80, 125, 42}
\definecolor{gommegoutte}{RGB}{228, 155, 15}
\definecolor{yellowL}{RGB}{247, 246, 26}
\definecolor{amber}{RGB}{255, 191, 0}
\definecolor{lemon}{RGB}{255, 216, 0}
\definecolor{carrotL}{RGB}{255, 163, 67}
\definecolor{cyanL}{RGB}{0, 204, 204} 
\definecolor{thrush}{RGB}{31, 206, 203} 
\definecolor{cyanD}{RGB}{1, 121, 111}

\DeclareMathOperator{\topwords}{top}

\DeclareMathOperator{\PMI}{PMI}

\DeclareMathOperator{\KL}{KL}
\DeclareMathOperator*{\norm}{norm}
\DeclareMathOperator{\PPL}{ppl}
\DeclareMathOperator{\Coh}{coh}
\DeclareMathOperator{\Cohtok}{coh_{topk}}
\DeclareMathOperator{\Cohintra}{coh_{intra}}
\DeclareMathOperator{\Div}{div}
\DeclareMathOperator{\Tplus}{T_{+}}  
\DeclareMathOperator{\Tplustoptok}{\tau_{+}^{topk}}

\DeclareMathOperator{\Tplustoptokatintra}{\tau_{+}^{topk@intra}}

\DeclareMathOperator{\df}{df}

\newcommand{\Tminus}{\mathop{\mathrm T_{-}}}

\newcommand*{\hm}[1]{#1\nobreak\discretionary{}{\hbox{$\mathsurround=0pt #1$}}{}}

\begin{document}
\title{Iterative Improvement of an Additively Regularized Topic Model}
%
%

\author{Alex Gorbulev\inst{1} \and
    Vasiliy Alekseev\inst{1}\orcidID{0000-0001-7930-3650} \and
    Konstantin Vorontsov\inst{2}\orcidID{0000-0002-4244-4270}
}

\authorrunning{A. Gorbulev et al.}

\institute{
    Moscow Institute of Physics and Technology, Dolgoprudny, Russia\\
    \email{\{gorbulev.ai,vasiliy.alekseyev\}@phystech.edu} \and
    Lomonosov Moscow State University, Moscow, Russia\\
    \email{k.vorontsov@iai.msu.ru}
}


\maketitle              
\begin{abstract}
    Topic modelling is fundamentally a soft clustering problem (of known objects---documents, over unknown clusters---topics).
    That is, the task is incorrectly posed.
    In particular, the topic models are unstable and incomplete.
    All this leads to the fact that the process of finding a good topic model (repeated hyperparameter selection, model training, and topic quality assessment) can be particularly long and labor-intensive.
    We aim to simplify the process, to make it more deterministic and provable.
    To this end, we present a method for iterative training of a topic model.
    The essence of the method is that a series of related topic models are trained so that each subsequent model is at least as good as the previous one, i.e., that it retains all the good topics found earlier.
    The connection between the models is achieved by additive regularization.
    The result of this iterative training is the last topic model in the series, which we call the iteratively updated additively regularized topic model (ITAR).
    Experiments conducted on several collections of natural language texts show that the proposed ITAR model performs better than other popular topic models (LDA, ARTM, BERTopic), its topics are diverse, and its perplexity (ability to ``explain'' the underlying data) is moderate.
    
\end{abstract}

\section{Introduction}

Topic modelling is the developing \cite{10031921} method of text analysis, which is used in sociological studies \cite{DIMAGGIO2013570}.
It is assumed that the text contains a set of \emph{hidden} topics that the topic model should find.
In probabilistic topic modelling, topics are represented as probability distributions on a set of words.
In addition to searching for the topics themselves, the topic model helps to assess the probability of each document belonging to each of the topics obtained.
It was probabilistic topic modelling that was used in the study of the dissemination of information about the COVID-19 pandemic in Croatia~\cite{pandemic2021}, and media coverage of climate change in Lithuania~\cite{climate2021}.
In addition to analyzing natural language texts, topic modelling can also be used in other applications, for example, for analyzing bank transactions~\cite{egorov2019topic}.

However, the topic modelling task has infinitely many solutions due to its incorrect formulation~\cite{vorontsov2015bigartm}.
In order to limit the number of solutions, regularizers are introduced.
Furthermore, regularizers can be used to obtain topic models with desired properties.
For example, a sparsing regularizer helps ensure that the probability distribution of a topic is concentrated in a small number of words, rather than spread throughout the entire vocabulary; the decorrelation regularizer helps to obtain more distinctive topics~\cite{artm2}.
Still, not all topics obtained by the topic model are good (satisfying a certain criterion, in general~---~interpretable).
In addition to good ones, topics can be bad (for example, when the most likely words of the topic are actually not related to each other from the point of view of a person, or when among the most common words of the topic there are ``stop words'' (service words) or ``background'' words (which do not carry any meaning other than ``lubricant'' for language)).
One can also identify a group of ``unremarkable'' topics: those that are neither good nor bad; topics that the researcher would not mind losing (for example, duplicates of already found good topics).

The natural incorrectness of the topic modelling problem, the resulting incompleteness and instability of topic models~\cite{alekseev2021topicbank,sukhareva2019postroenie}, and the presence of bad topics among those found by the model lead to the fact that experiments with topic models can take a long time and be haphazard.
Model hyperparameters selection, training, model quality evaluation, model topic analysis (automatic or semi-automatic).
If the model is not satisfactory, the process is repeated.
In this case, the good topics that have been found are \emph{lost}.
So it can take a long time until suitable hyperparameters are found.

The main idea of this paper is to build a clear and systematic path from an initial topic model to a good one.
Without avoiding the need to train several models, we propose to train models in a \emph{connected} manner, one after another.
So that each successive model \emph{fixes} all the good topics found earlier, and \emph{filters} out the bad ones, i.e., that the remaining topics do not include the bad topics found earlier.
Throughout the paper, we will often call this sequence of topic models trained one after another (and, depending on the context, the most recent, final, model in the sequence) as just ``the iterative topic model''.
Thus, in this context, ``iteration''~---~is the training of a single model.

Additive regularization of topic models (ARTM) is used to solve the problem of fixing good topics and filtering out bad ones.
The ARTM approach helps to optimize models by the sum of several criteria~\cite{artm}, which helps to take into account the peculiarities of the text collection and limit the number of solutions to the topic modelling problem.

The main contribution of the paper can be summarized as follows:
\begin{itemize}
  \item An iteratively updated additively regularized topic model is presented.
  \item Regularizers for topic fixation and filtering are introduced.
  \item Experiments comparing the proposed model with several other topic models on several natural language collections show that the iterative topic model is able to accumulate the highest percentage of good topics, whereas the model topics are also different.  
\end{itemize}

\section{Related Work}

\subsubsection{Probabilistic Topic Modelling}

In the paper~\cite{hofmann1999probabilistic} the simplest topic model PLSA is presented, which solves the problem of matrix decomposition of the known matrix of words-in-documents frequencies as the product of the matrices of words-in-topics probabilities~$\Phi$ and topics-in-documents probabilities~$\Theta$ without any additional constraints, except that the columns of the probability matrices should be stochastic.
The authors~\cite{blei2003latent} proposed the LDA model, which later became very popular, where the same matrix decomposition problem is solved, but with an additional restriction on the columns of the probability matrices of words-in-topics and topics-in-documents: they are assumed to be generated by the Dirichlet distribution.

\subsubsection{Additive Regularization of Topic Models}

The paper~\cite{vorontsov2015bigartm} is a starting point in the development of the theory of additive regularization of topic models, in which our work fits into.
Both PLSA and LDA models can be realized within the ARTM approach.
In addition, regularizers provide a convenient tool to obtain topic models with desired properties, such as topic sparsity, topic distinctness, and division of all topics into subject and background ones.
In~\cite{irkhin2020additive}, a topic model that learns without a probability matrix of topics-in-documents is proposed; topics of this model are immediately (without additional regularization) sparse.

\subsubsection{Neural Topic Models}


Currently, much attention has been paid to the possibility of using neural networks for topic modelling~\cite{dai2023llm}: neural network-based topic models have been proposed~\cite{wang2020neural,grootendorst2022bertopic,rahimi2024antm}, large language models have also been used to evaluate the quality of the resulting topics~\cite{yang2024llm}.

The neural topic model against which we are going to compare the probabilistic topic models implemented within the ARTM framework is BERTopic~\cite{grootendorst2022bertopic}, which uses word embeddings derived from the BERT model~\cite{devlin2018bert}.
The BERTopic model consists essentially of several ``blocks''.
The first block creates embeddings of documents using a neural embedding model.
The next block uses UMAP~\cite{mcinnes2018umap} to reduce the dimensionality of these embeddings before clustering.
Then the actual document clustering takes place using HDBSCAN~\cite{mcinnes2017hdbscan} algorithm.
Finally, using TF-IDF \cite{sparck1972statistical}, where the previously obtained clusters---groups of documents of the same topic are used as ``documents'', BERTopic determines the top words of each topic.

\subsubsection{Intrinsic Topic Model Quality Measures}

One popular measure of the quality of a topic model as a whole is \emph{perplexity}~\cite{jelinek1977perplexity,vorontsov2015bigartm}.
Sometimes, perplexity is even used to determine the ``optimal'' number of topics in a text collection~\cite{griffiths2004finding}.
Perplexity is closely related to the plausibility of a collection $\mathcal (\Phi, \Theta)$ and is expressed by the following formula:
\begin{equation}\label{eq:perplexity}
  \PPL(\Phi, \Theta) = e^{-\mathcal L(\Phi, \Theta)}
\end{equation}
The smaller the perplexity, the better the model ``fits'' the data.

The next quality criterion, which we consider particularly important and will use in this paper, is the \emph{diversity} of topics.
It is believed that in a good topic model, topics should be different~\cite{vorontsov2015additive,cao2009density}.
In this work, we use Jensen--Shannon divergence~\cite{deveaud2014accurate} as a metric to evaluate the dissimilarity of topics:
\begin{equation}\label{eq:js-div}
  \Div(\Phi) = \frac{1}{\binom{T}{2}} \sum\limits_{\substack{\phi_i \not= \phi_j \\ \phi_i, \phi_j \in \Phi}} \sqrt{\frac{1}{2} \left(\vphantom{\frac{1}{2}} \KL(\phi_i \parallel \phi_j) + \KL(\phi_j \parallel \phi_i)\right)}
\end{equation}  

In \cite{newman2010automatic,mimno2011optimizing,lau2014machine} works, authors propose a way of assessing topic quality called topic \emph{coherence}: where a decision about topic quality is made based on how often word pairs of the most frequent topic words occur near each other in the text (compared to the number of times one and the other word do not necessarily occur near each other in the text).
This topic coherence, based on top-word co-occurences, is expressed as follows:
\begin{equation}\label{eq:coh}
    \Cohtok(t) = \frac{1}{\binom{k}{2}} \sum\limits_{\substack{w_i \not= w_j \\ w_i, w_j \in \topwords_k(t)}} \PMI(w_i, w_j),\quad
    \PMI(w_i, w_j) = \log_2 \frac{p(w_i, w_j)}{p(w_i) p(w_j)}  
\end{equation}
where $p(w_i, w_j)$, $p(w_i)$ are the probabilities of encountering one word $w_i$ or two words $w_i, w_j$ from the top words of the topic $\topwords_k(t)$ together in a window of some size in the text.
The probabilities are estimated using the known frequencies of the words in the documents.

In the paper \cite{intracoh}, another approach called \emph{intra-text coherence}, is proposed to evaluate the quality of topics.
The authors hypothesize about the segmental structure of texts~\cite{skachkov2018improving}, which states that the words of topics occur in text not randomly, but close to each other, in groups, or \emph{segments}.
Thus, if a topic is good, it is consistent with the segment structure hypothesis, and the average \emph{segment length} of this topic will be greater than the length of word segments of bad topics.
The intra-text coherence $\Cohintra$ just estimates the average length of a text segment consisting of words of the topic under study.
This coherence method is not expressed as a formula, but rather as an algorithm that results in the entire collection being viewed from beginning to end.
This, in particular, makes intra-text coherence more computationally expensive than top-word coherence (for a quick computation of which it is enough to go through the collection once in advance to make a matrix of size $W \hm\times W$ of word co-occurrences).

\subsubsection{Iterative Approach to Topic Modelling}

Topic model training through iterative updates is not something new.
Thus, the solution to the probabilistic topic modelling problem itself is an iterative algorithm.
(Although, in this context, ``iteration'' means just one update of $\Phi$ and $\Theta$ matrices.)
In addition, in applications, topic models can be used to analyze collections that change over time: for example, news streams, databases of scientific articles~\cite{gerasimenko2023incremental}---where an already trained topic model needs to be updated on newly arrived data.

This paper, on the other hand, concentrates on updating the topic model with a static collection.
So, it is in a sense a ``wrapper'' over the process of training a topic model.
In this respect, this work is closest to \cite{sukhareva2019postroenie,alekseev2021topicbank}.
Thus, in \cite{alekseev2021topicbank}, the authors introduce the concept of TopicBank---a collection of good different topics that are accumulated in the process of multiple topic model training.
The authors propose to use TopicBank as a way to evaluate the quality of newly trained models~\cite{korenvcic2021topic}.
So, TopicBank does not have to be a good topic model, because its low perplexity as a model was not a criterion for accumulating new good topics.

\section{Method}

\subsection{Probabilistic Topic Modelling}

Let $D$ be a collection of texts, and $W$ be a set of terms.
Among the terms, there can be both words and word combinations~\cite{artm2}.
We represent each document $d \in D$ as a sequence of $n_d$ terms $\left(w_1, \ldots, w_{n_d} \right)$ from the set $W$~\cite{artm}.
The text collection is assumed to contain a finite set $T$ of \emph{hidden} topics.
The document collection $D$ is considered as a sample from a discrete distribution $p(d, w, t)$ on the finite set $D \hm\times W \hm\times T$~\cite{artm2}.
According to the formula of total probability and the conditional independence hypothesis (which states that a word refers to a topic regardless of which document the word occurs in), the distribution of terms in documents $p(w \mid d)$ is described by a probabilistic mixture of the distributions of terms-in-topics $\phi_{wt} \hm= p(w \hm\mid t)$ and topics-in-documents $\theta_{td} \hm= p (t \hm\mid d)$ as follows~\cite{vorontsov2015bigartm}:
\begin{equation}\label{eq:p(w|d)}
    p(w \mid d)
    = \sum \limits_{t \in T} p (w \mid t) p (t \mid d)
    = \sum \limits_{t \in T} \phi_{wt} \theta_{td}
\end{equation}
Thus, the probabilistic topic model~\eqref{eq:p(w|d)} describes how documents $D$ are generated by a mixture of distributions $\theta_{td}$ and $\phi_{wt}$.
The task of topic modelling is to find, given a collection of documents $D$, the parameters $\phi_{wt}$ and $\theta_{td}$ that approximate the frequency estimates of the conditional probabilities $\widehat{p} (w \mid d) \hm= n_{dw} \hm/ n_d$ known from the text ($n_{dw}$ is a number of occurrences of the word $w$ in the document $d$).
Since $|T|$ is usually much smaller than $|W|$ and $|D|$, this reduces to the problem of low-rank stochastic matrix decomposition~\cite{artm2}: $F \hm\approx \Phi \Theta$, where $F \hm= {(\widehat{p}_{dw})}_{|W| \times |D|}$ is the words-in-documents frequency matrix, $\Phi \hm= {(\phi_{wt})}_{|W| \times |T|}$ is the words-in-topics probability matrix, and $\Theta \hm= {(\theta_{td})}_{|T| \times |D||}$ is the probability matrix of words-in-documents.
The matrices $F$, $\Phi$, and $\Theta$ are all stochastic: their columns $f_d$, $\phi_t$, and $\theta_d$, respectively, are non-negative, normalized, and represent discrete probability distributions.
\emph{Topic} $t$ in probabilistic topic modelling usually is associated with its corresponding column $\phi_t$ (although the row $(\theta_{td})_{d \in D}$ also characterizes the topic).

The paper~\cite{hofmann1999probabilistic} presents one of the earliest, yet one of the simplest and most straightforward, topic models---the PLSA model, where distributions~\eqref{eq:p(w|d)} are trained by maximizing the log-likelihood of a collection with linear constraints.

\emph{Likelihood} is the probability of the observed data as a function of the parameters $\Phi$ and $\Theta$:
\[
    p(\Phi, \Theta)
    = \prod_{d \in D} \prod_{w \in d} p(d, w)^{n_{dw}}
    = \prod_{d \in D} \prod_{w \in d} p(w \mid d)^{n_{dw}} p(d)^{n_{dw}} \to \max_{\Phi, \Theta}
\]

Maximizing the logarithm of the likelihood $\log p(\Phi, \Theta)$ is equivalent to the following:
\begin{equation}\label{eq:logL}
    \mathcal L(\Phi, \Theta)
    = \sum_{d\in D} \sum_{w \in d} n_{dw} \ln \sum_{t\in T} \phi_{wt} \theta_{td}
    \to \max_{\Phi,\Theta}
\end{equation}

\subsection{Additive Regularization of Topic Models}\label{sec:artm}

ARTM is based on maximizing \emph{regularized} log-likelihood~\eqref{eq:logL}:
\begin{equation}\label{eq:logL-reg}
 \mathcal L(\Phi, \Theta) + R(\Phi, \Theta) \to \max\limits_{\Phi, \Theta}
\end{equation}
where $R(\Phi, \Theta) \hm= \sum_{i=1}^n \tau_i R_i(\Phi, \Theta)$ is a weighted sum of regularizers $R_i(\Phi, \Theta)$ with weights $\tau \hm\in \mathbb R$.
Thus, regularizers are additives to the function being optimized, imposing additional constraints and, at the same time, leading to final topics satisfying additional properties~\cite{vorontsov2015bigartm}.
For example, if the collection $D$ is unbalanced, one can require the model to have topics of different sizes~\cite{veselova2020topic}.
The point of local extremum of the problem~\eqref{eq:logL-reg} satisfies a system of equations that can be solved by an iterative method equivalent to the EM algorithm, updating $\Phi$ and $\Theta$ at each iteration \cite{vorontsov2014tutorial,vorontsov2015bigartm,vorontsov2015additive}:
\[
  \left\{
    \begin{aligned}
      p_{tdw} &= \norm_{t \in T} (\phi_{wt} \theta_{td})\\
          \phi_{wt} &= \norm_{w \in W} \left(n_{wt} + \phi_{wt} \frac{\partial R}{\partial \phi_{wt}} \right)\\
          \theta_{td} &= \norm_{t \in T} \left(n_{td} + \theta_{td} \frac{\partial R}{\partial \theta_{td}} \right)
    \end{aligned}
  \right.
\]
where $p_{tdw} \hm= p(t \hm\mid d, w)$, $\norm$ is vector normalization operator ($\norm_{w \hm\in W} n_{dw} \hm= f_d$), $n_{wt} \hm= \sum_{d \in D} n_{dw} p_{tdw}$, $n_{td} \hm= \sum_{w \in d} n_{dw} p_{tdw}$.
It can be seen that each new regularizer~\eqref{eq:logL-reg} is eventually expressed as an additive at M-step.

Let us introduce a couple of popular ARTM regularizers which also play an important role in this paper.

\paragraph{Smoothing and Sparsing Regularizers} Topics $T$ of a topic model can be divided into topics of two types: domain-specific $S$ and background ones $B$. 
Subject topics consist of specialized words and are assumed to be sparse and loosely correlated.
Background topics, on the other hand, consist of general vocabulary words and are evenly distributed throughout the documents in the collection.

The idea of a smoothing (sparsing) regularizer is to make the distributions $\phi_t$ and $\theta_d$ close to (far from) the uniform distributions $\beta_t$ and $\alpha_d$:
\begin{equation}\label{eq:smooth-sparse}
    R(\Phi, \Theta) = \beta_0 \sum\limits_{t \in H}\sum\limits_{w \in W} \beta_{wt} \ln \phi_{wt}
    + \alpha_0 \sum\limits_{t \in H}\sum\limits_{d \in D} \alpha_{td} \ln \theta_{td} \to \max_{\Phi, \Theta}
\end{equation}
where, for smoothing, $H \hm\equiv B$ and $\beta_0, \alpha_0 \hm> 0$; and, for sparsing, $H \hm\equiv S$ and $\beta_0, \alpha_0 \hm< 0$.


\paragraph{Decorrelation Regularizer} The desirable property of topics of a topic model is that they are different~\cite{vorontsov2015additive}.
The decorrelation regularizer increases the Euclidean distance between topic columns:
\begin{equation}\label{eq:decorrelation}
  R(\Phi) = -\frac{\tau}{2} \sum\limits_{t \in T} \sum\limits_{s \in T \setminus t} \sum\limits_{w \in W} \phi_{wt} \phi_{ws} \to \max_{\Phi}
\end{equation}

The regularizer additive on the M-step will be as follows:
\[
  \phi_{wt} \frac{\partial R}{\partial \phi_{wt}} = -\tau \phi_{wt} \sum\limits_{s \in T \setminus t} \phi_{ws}
\]

\subsection{Iterative Additively Regularized Topic Model}

\begin{figure}[!ht]
    \centering
    \includegraphics[width=0.6\linewidth]{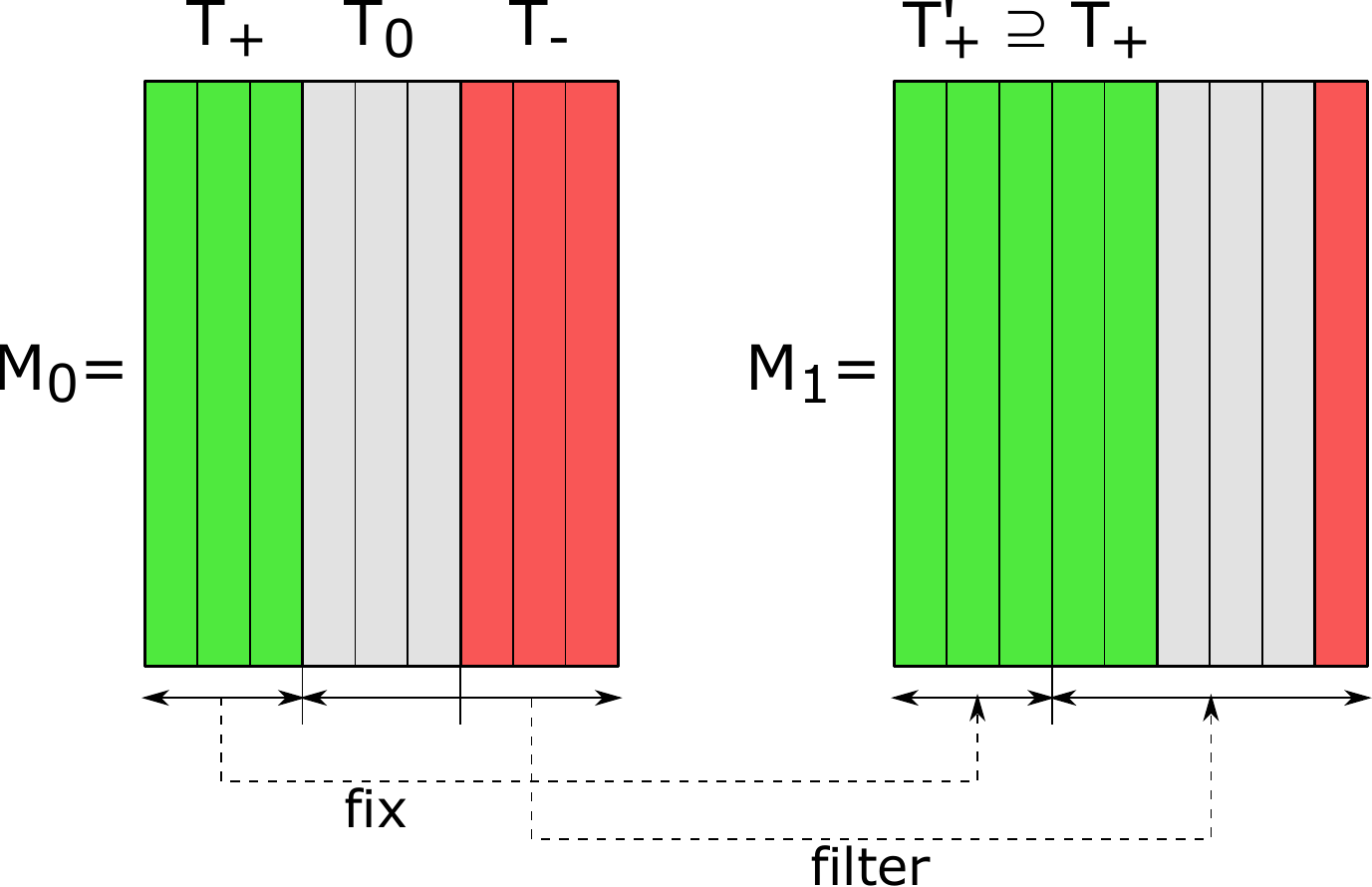}
    \caption{%
        The idea of an iterative approach to topic model improvement. The topics of the initial model $M_0$ are automatically or semi-automatically classified into good $T_+$, bad $T_-$, and ``unremarkable'' $T_0$ (those that you can't afford to lose, not bad, but not relevant for the study, for example, they can be duplicates of topics from $T_+$). Next, a new topic model $M_1$ is trained so that it retains all the topics of $T_+$, and at the same time has no topics from $T_-$. Thus, model $M_1$ is at least as good as model $M_0$ in terms of the number of good topics $T_+'$, and possibly even better: $T_+' \hm\supseteq T_+$.
    }
    \label{fig:itar-scheme}
\end{figure}

The main idea of the proposed iteratively updated model is that, given one model, train the next model so that it is guaranteed to be at least as good as the previous model (see Fig.~\ref{fig:itar-scheme}).
This is achieved by using two regularizers: topic fixation and topic filtering (decorrelation with good $T_+$ and bad $T_-$ topics collected previously, whose columns are collected into $\widetilde{\Phi}$ matrix):
\begin{equation}\label{eq:itar}
\begin{split}
  &\overbrace{\mathcal L(\Phi, \Theta) + R_{\mathrm{sparse}}(\Phi) + R_{\mathrm{decorr}}(\Phi)}^{\mathrm{ARTM}} \\
  &\underbrace{
    \hphantom{\mathcal L(\Phi, \Theta)\,}+ R_{\mathrm{fix}}(\Phi, \widetilde{\Phi}) + R_{\substack{\mathrm{filter}\\\mathrm{bad}}}(\Phi, \widetilde{\Phi}) + R_{\substack{\mathrm{filter}\\\mathrm{good}}}(\Phi, \widetilde{\Phi})
    }_{\mathrm{ITAR}}\to \max\limits_{\Phi, \Theta}
\end{split}
\end{equation}

The topic fixing regularizer acts like the smoothing one~\eqref{eq:smooth-sparse}, only now instead of the uniform distribution (KL divergence with which the regularizer tends to reduce for the selected topics), it is the one we want to keep:
\begin{equation}\label{eq:r-fix}
  R_{\mathrm{fix}}(\Phi, \widetilde{\Phi})|_{\textcolor{DarkOrchid}{\tau \gg 1}} = \tau \sum\limits_{t \in T_+} \sum\limits_{w \in W} \widetilde{\phi}_{wt} \ln \phi_{wt} \to \max\limits_{\Phi}
\end{equation}


The topic filtering regularizer acts like the decorrelation one~\eqref{eq:decorrelation}, only now the topics of the trained model are decorrelated not with each other, but with the topics collected previously:
\begin{equation}\label{eq:r-decorr1}
    R_{\substack{\mathrm{filter}\\\mathrm{\textcolor{Bittersweet}{bad}/\textcolor{PineGreen}{good}}}}(\Phi, \widetilde{\Phi})|_{\tau > 0} = -\tau \sum \limits_{t \in T'} \sum \limits_{s \in \textcolor{Bittersweet}{T_-}\textcolor{PineGreen}{/T_+}} \sum \limits_{w \in W} \phi_{wt} \widetilde{\phi}_{ws} \to \max\limits_{\Phi}
\end{equation}
where $T' \hm= T \hm\setminus T_+$ are free topics in the new model (which we do not fix).
We propose to correlate with both bad and good topics that are collected from previous model training iterations.
The idea of decorrelation with bad topics seems clear: we do not want bad topics already seen before to reappear in the new topic model.
Decorrelation with the good ones makes sense so that the model does not try to find the same good topics again (we want different good topics).

Additive on M-step from a regularizer filtering out bad topics:
\[
  \phi_{wt} \frac{\partial}{\partial \phi_{wt}} R_{\substack{\mathrm{filter}\\\mathrm{bad}}} = -\tau [t \in T'] \phi_{wt} \sum\limits_{s \in T^{-}} \widetilde \phi_{ws}
\]
where the expression $[t \hm\in T']$ just means a Boolean indicator, that is, $[t \hm\in T'] \hm= (1 \mbox{ if } t \hm\in T' \mbox{ else } 0)$.

It can be seen that the effect of such a regularizer is actually the decorrelation with just the ``average bad'' topic $\sum\limits_{s \in T^{-}} \widetilde \phi_{ws}$!
Although the original idea was to have a new topic trained unlike \emph{any} of the collected bad ones.
We therefore present another version of the topic filtering regularizer, which aims to correct this issue:
\begin{equation}\label{eq:r-decorr2}
    R_{\substack{\mathrm{filter2}\\\mathrm{\textcolor{Bittersweet}{bad}/\textcolor{PineGreen}{good}}}}(\Phi, \widetilde{\Phi})|_{\tau > 0} = -\frac{\tau}{2} \sum\limits_{t \in T'} \sum\limits_{s \in \textcolor{Bittersweet}{T_-}\textcolor{PineGreen}{/T_+}} \left(\sum\limits_{w \in W} \phi_{wt} \widetilde \phi_{ws}\right)^2 \to \max_{\Phi}
\end{equation}
\[
  \phi_{wt} \frac{\partial}{\partial \phi_{wt}} R_{\substack{\mathrm{filter2}\\\mathrm{bad}}} = -\tau [t \in T'] \phi_{wt} \sum\limits_{s \in T_-} \widetilde \phi_{ws} \sum\limits_{u \in W} \phi_{ut} \widetilde \phi_{us}
\]

It can be seen that the values calculated by the second version of the regularizer~\eqref{eq:r-decorr2} are by orders of magnitude smaller than the values of the filtering regularizer of the first version~\eqref{eq:r-decorr1} (this means that for the same effect on the model, the coefficient $\tau$ of the second regularizer should be orders of magnitude larger than $\tau$ of the first one).

We will denote the iterative model using the first version~\eqref{eq:r-decorr1} of the filtering regularizer as \emph{ITAR}, and the model with the second version~\eqref{eq:r-decorr2} of the regularizer as \emph{ITAR2}.

\section{Experiment}

\subsection{Methodology}\label{sec:methodology}

In the experimental part, we want to verify the following points:
\begin{itemize}
  \item is the number of good topics in the proposed model really increases iteratively?
  \item does the iterative model outperform other topic models by the final number of good topics?
\end{itemize}

The methodology is as follows.
We take several topic models for comparison.
They are supposed to be compared by the number of good topics on several collections of natural language texts.
In separate experiments, we will train models with different numbers of topics: $T \hm= 20$ and $T \hm= 50$ (a good topic model could be trained with an arbitrary number of topics~\cite{bulatov2024optnum}).
For each model, several (namely $20$) trainings with different model initializations are performed.
Iterative models are updated from iteration to iteration, while the other models are trained completely anew at each iteration.
The final iterative model is the model at the last iteration, the final non-iterative model is the best model in terms of the number of good topics from the whole series.

Topics with high coherence {\cite{newman2010automatic,intracoh} are considered good.
The high coherence threshold is also found experimentally by percentile analysis of all topics obtained from all training iterations of all ARTM-based non-iterative probabilistic topic models under study ($80\%$ is taken as the coherence threshold to consider a topic good; $20\%$ is a threshold to consider a topic bad; for each dataset and for each number of topics in the model, the absolute coherence thresholds were different).
In this work, we use a document~\cite{mimno2011optimizing} as the co-occurrence window for the coherence calculation~\eqref{eq:coh}; the number of top words is $k \hm= 20$; also, instead of ``plain PMI''~\cite{fano1968transmission,church1990word} we use its positive~\cite{dagan1993contextual} version:
\[
  \PMI_+(w_i, w_j) = \max\left(\ln \frac{p(w_i, w_j)}{p(w_i) p(w_j)}, 0\right)
\]

The experiments were performed in Python using the open source libraries TopicNet\footnote{{\UrlFont\url{https://github.com/machine-intelligence-laboratory/TopicNet}}.}~\cite{bulatov2020topicnet} and BigARTM\footnote{{\UrlFont\url{https://github.com/bigartm/bigartm}}.}~\cite{vorontsov2015bigartm}.
The source code of the proposed iterative topic model, as well as the source code and the results of the experiments conducted, are also publicly available.\footnote{{\UrlFont\url{https://github.com/machine-intelligence-laboratory/OptimalNumberOfTopics}}.}

\subsection{Data}\label{sec:data}

Several text collections are used: some in Russian, some in English~(see Tab.~\ref{tab:datasets}).
All datasets have already been preprocessed specifically for topic modelling (modalization, lemmatization, ngram extraction, removal of stop words).
20Newsgroups\footnote{{\UrlFont\url{http://qwone.com/~jason/20Newsgroups}}.} is a popular dataset in topic modelling.  
PostNauka dataset was first used in~\cite{intracoh,belyy2018quality},  
RuWiki-Good and ICD-10\footnote{{\UrlFont\url{https://en.wikipedia.org/wiki/ICD-10}}.} datasets were collected by the authors of~\cite{bulatov2020topicnet}.
RTL-Wiki-Person was first used in~\cite{chang2009reading,roder2015exploring}.

The only preprocessing that we did with the datasets was vocabulary filtering.
Thus, the datasets were typically multimodal (e.g., plain text, bigrams, author, title), but only one main modality (plain text) was used in the experiments.
In addition to modality filtering, token filtering by frequency was applied: $\df_{\min} \hm= 5$, $\df_{\max} \hm= 0.5$ for RuWiki-Good; $\df_{\min} \hm= 2$, $\df_{\max} \hm= 0.5$ for RTL-Wiki-Person and for ICD-10 (where $\df$ means the frequency of token occurrence in the documents of the collection, absolute (how many documents) or relative (proportion of documents)).

All datasets used are in the public domain.\footnote{{\UrlFont\href{https://huggingface.co/TopicNet}{https://huggingface.co/TopicNet}}.}

\begin{table}[!ht]
    \centering
    \caption{%
        Datasets used in the experiments (D means number of documents, Len represents average document length, W means vocabulary size (after filtering out very rare and very frequent words), Lang is language, BOW is an indicator of the presence of text in Bag-of-Words format, NWO is an indicator of availability of text with natural word order).
        20Newsgroups dataset includes the train split only.
    }
    \label{tab:datasets}
    \begin{tabular}{l|ccc|c|cc}
        \toprule
        \textbf{Dataset} & \textbf{D} & \textbf{Len} & \textbf{W} & \textbf{Lang} & \textbf{BOW} & \textbf{NWO} \\
        \midrule
        PostNauka       & 3404  & 421  & 19186 & Ru & $\checkmark$ & $\checkmark$ \\
        20Newsgroups    & 11301 & 93   & 52744 & En & $\checkmark$ & $\checkmark$ \\
        RuWiki-Good     & 8603  & 1934 & 61688 & Ru & $\checkmark$ & {} \\
        RTL-Wiki-Person & 1201  & 1600 & 37739 & En & $\checkmark$ & {} \\
        ICD-10          & 2036  & 550  & 22608 & Ru & {}            & $\checkmark$ \\
        \bottomrule
    \end{tabular}
\end{table}

\subsection{Models}

The following topic models are used for comparison with the proposed iterative topic model~\eqref{eq:itar}.
\emph{PLSA}, a model with a single hyperparameter $T$~\cite{hofmann1999probabilistic};
\emph{LDA}, whose columns $\Phi$ and $\Theta$ are generated by Dirichlet distributions~\cite{blei2003latent};
\emph{Sparse}, a model with additive regularization~\cite{vorontsov2015bigartm} consisting of topic sparsing and smoothing (which is applied to one additional background topic);
\emph{Decorr}, a model with additive regularization, consisting of topic decorrelation~\eqref{eq:decorrelation} and smoothing~\eqref{eq:smooth-sparse} (again, smoothing is applied for one additional background topic only);  
\emph{TLESS}, a model without $\Theta$ matrix, with inherently sparse topics~\cite{irkhin2020additive};
\emph{BERTopic}, a neural topic model \cite{grootendorst2022bertopic};
\emph{TopicBank}, an iteratively updated topic model but without regularizers~\cite{alekseev2021topicbank}.

All models except BERTopic are implemented within the TopicNet/ARTM framework.

A basic model that is trained with different initializations at ITAR model training iterations is the one with sparsing, smoothing, and decorrelation ARTM regularizers, and with additional fixing~\eqref{eq:r-fix} and filtering regularizers~\eqref{eq:itar} (for ITAR model it is defined by~\eqref{eq:r-decorr1}, for ITAR2 model---by~\eqref{eq:r-decorr2}).

All models except PLSA have one or more adjustable hyperparameters.
The hyperparameters of the regularized models (Sparse, Decorr, ITAR, ITAR2) are basically the regularization coefficients~$\tau$.
The regularization coefficients of the Sparse and Decorr models were used relative and searched on a grid: $\{-0.05, -0.1\}$ for the sparse regularizer, $\{0.05, 0.1\}$ for the smooth one, and $\{0.01, 0.02, 0.05, 0.1\}$ for the decorrelation regularizer (coefficients leading to the model with minimal perplexity were chosen).  
The regularization coefficient for the fixing regularizer of the ITAR and ITAR2 models was set equal to the absolute value of $10^9$ for all datasets except RuWiki-Good (and for it the coefficient was set equal to $10^{12}$).
For the filtering regularizer, the regularization coefficients were also absolute, and searched on the grid: $\{10, 100, \ldots, 10^{10}\}$ for the ITAR's regularizer~\eqref{eq:r-decorr1}, $\{10, 100, \ldots, 10^{12}\}$ for the ITAR2's one~\eqref{eq:r-decorr2} (the minimum coefficient was chosen, which led to a noticeable, but not very large (about $10\%$), deterioration of perplexity).
As a result, the following regularization coefficients were fixed: $\tau \hm= -0.05$ for sparsing; $\tau \hm= 0.05$ for smoothing; $\tau \hm= 0.01$ for decorrelation; $\tau \hm= 10^5$ for ITAR's filtering for PostNauka, 20Newsgroups, ICD-10 datasets, and $\tau \hm= 10^6$ for RuWiki-Good and RTL-Wiki-Person datasets; $\tau \hm= 10^8$ for the ITAR2's filtering for PostNauka, 20Newsgroups, ICD-10 datasets, $\tau \hm= 10^{10}$ for RuWiki-Good, and $\tau \hm= 10^9$ for RTL-Wiki-Person.
For the LDA model, we used symmetric priors (we also compared it to asymmetric~\cite{wallach2009rethinking} ones and found little difference; ``heuristic'' priors~\cite{rosen2016mobile} led to a somewhat higher perplexity value).

The BERTopic model differs in that there is no possibility to explicitly set the desired number of topics.
However, the number of topics was selected close to the desired one ($20$ or $50$) with the help of HDBSCAN's flat clustering submodule\footnote{{\UrlFont\href{https://github.com/scikit-learn-contrib/hdbscan/pull/398}{https://github.com/scikit-learn-contrib/hdbscan/pull/398}}.}.
Thus, when initializing the BERTopic model, basically only $\epsilon$ parameter\footnote{\texttt{cluster\_selection\_epsilon}.} (which was the one responsible for the number of topics in the model) was selected and set.
The rest of the hyperparameter values were default values.

The TopicBank model is used in two versions: \emph{TopicBank}, as in the original paper~\cite{alekseev2021topicbank}; and \emph{TopicBank2}, when multiple training of ARTM-regularized models (sparsing, smoothing and decorrelation) rather than PLSA ones are used to create the TopicBank.
In addition to changing the base model, the coherence threshold for selecting topics for the bank was changed for TopicBank2 model: instead of $90\%$ percentile of the topic coherence value of just one newly trained model~\cite{alekseev2021topicbank}, the same threshold was used as for collecting good topics for the ITAR model.  

At each iteration of training, a new \texttt{seed} equal to the iteration number was set in the model during initialization (which determines $\Phi$ matrix initialization for ARTM-based models, and UMAP behaviour for BERTopic).


\section{Results and Discussion}

\subsection{Top-Token Coherence}\label{sec:toptok-results}

Experiments have shown that the iterative model contains the most good topics.
More than $80\%$ of the topics of the iterative model (see Fig.~\ref{fig:toptok-plots}) can be good.
At the same time, its topics are different, and the perplexity of the whole model, although not the smallest among the considered models, is moderate (see the results for several datasets in~Tab.~\ref{tab:toptok-results}).
The fact that the perplexity of the iterative model is higher than the minimum is understandable, since it is a model trained with additional constraints (with regularization).
The minimum perplexity is possessed by the simplest PLSA and LDA models.

Regarding regularization, the following has also been observed.
Since the fixation of topics in the iterative model increasingly limits its freedom (as the number of good topics grows), there may come a point when there are so many good topics that the remaining free topics ``degenerate'' and become null.
To avoid this, the iterative model was trained with the stopping criterion by the number of already collected good topics: when the model was trained for $T \hm= 20$, it was considered that accumulation of at least $20 \hm- 2 \hm= 18$ of good topics was enough; when the model was trained for $T \hm= 50$, we considered $50 \hm- 5 \hm= 45$ of good topics sufficient for stopping (thus, $90\%$ of model topics in both cases).
However, in the iterative model, there still could be more than $90\%$ good topics, because more than one topic can be added at an iteration.

\begin{figure}[!ht]
    \centering
    \begin{subfigure}[b]{0.8\textwidth}
        \centering
        \includegraphics[width=\textwidth]{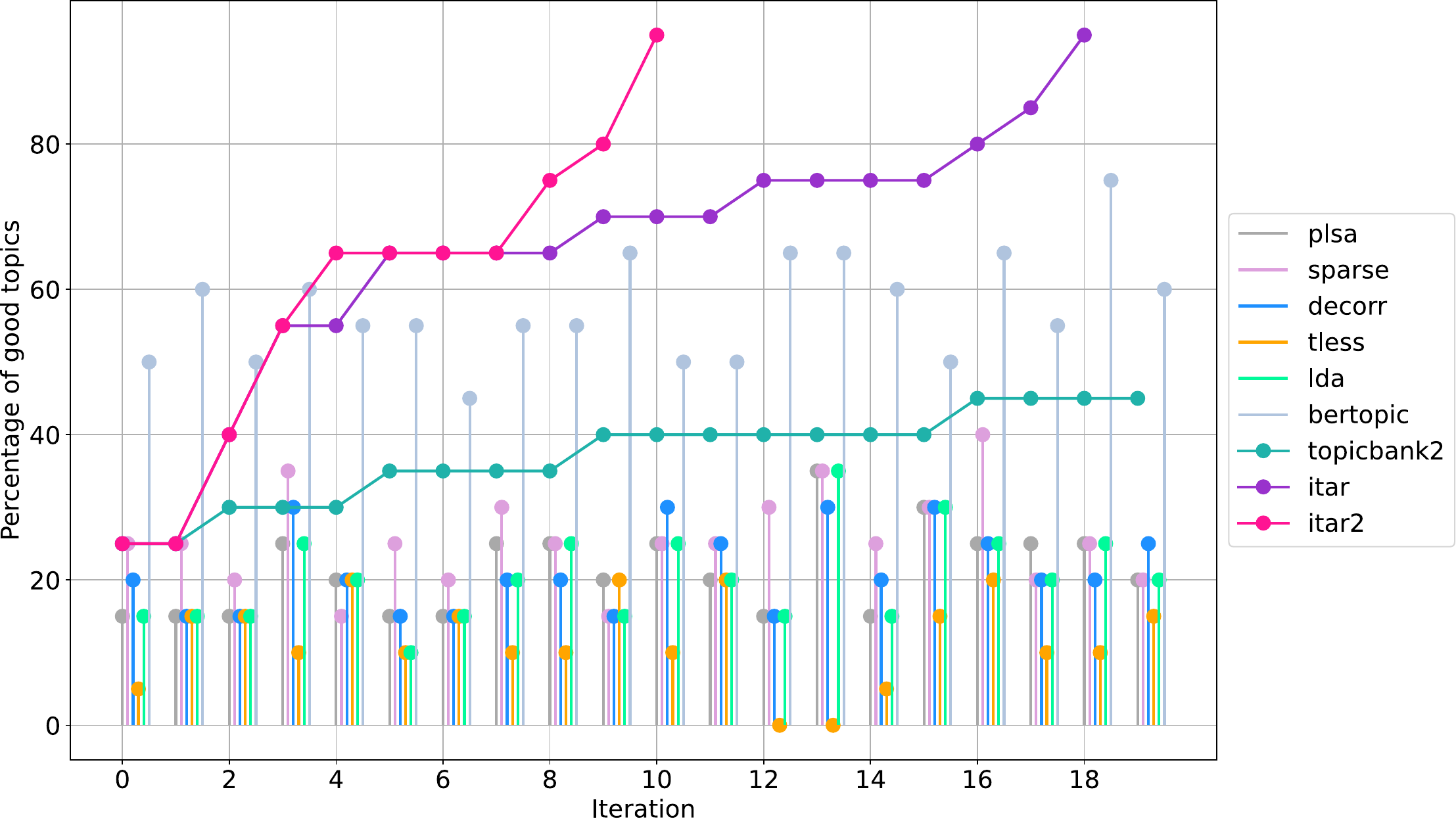}
        \caption{%
            RuWiki-Good, $T \hm= 20$.
        }
        \label{fig:toptok-plots-rwg}
    \end{subfigure}

    \begin{subfigure}[b]{0.8\textwidth}
        \centering
        \includegraphics[width=\textwidth]{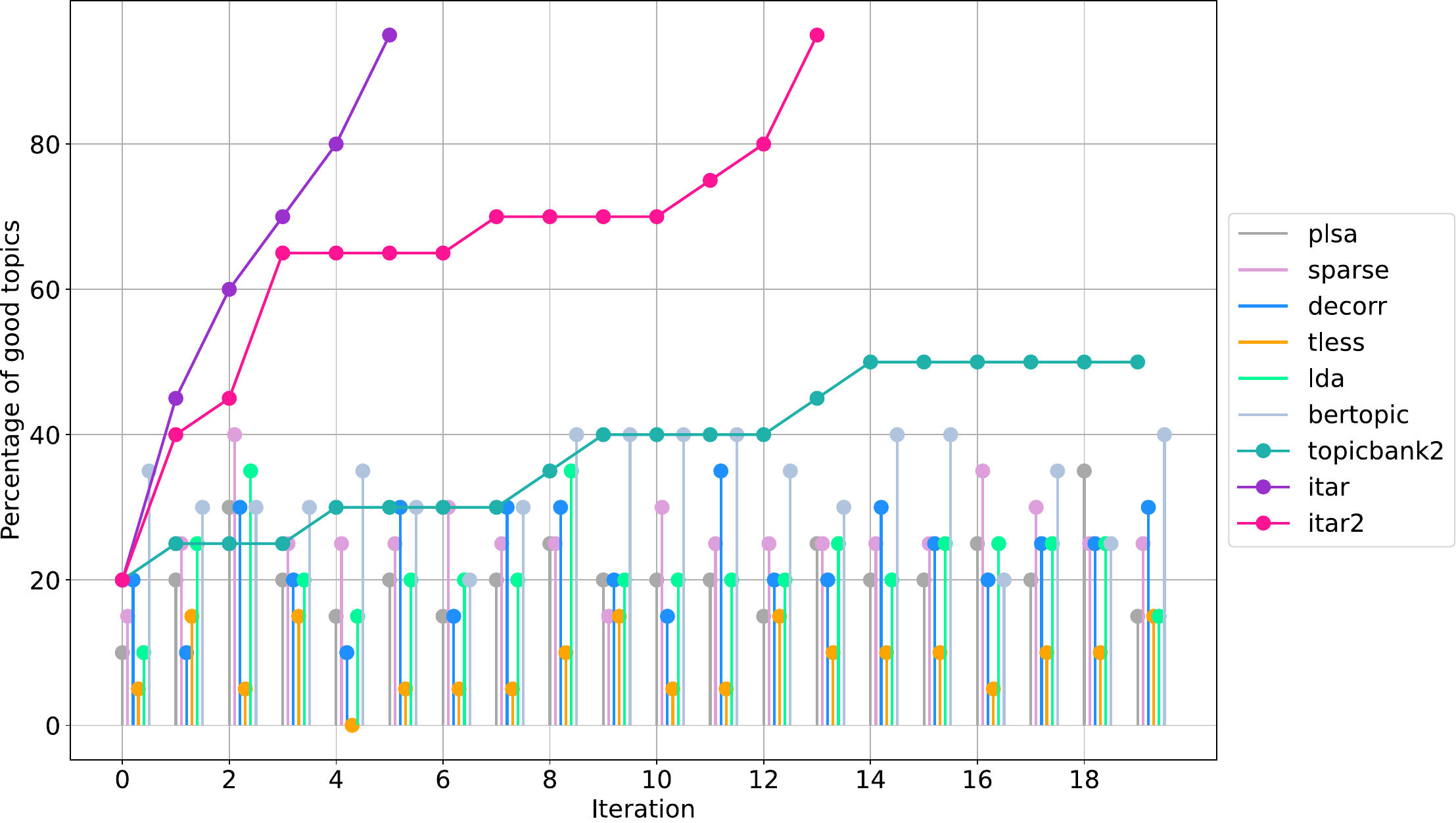}  
        \caption{%
            RTL-Wiki-Person, $T \hm= 20$.
        }
        \label{fig:toptok-plots-rtl-wiki}
    \end{subfigure}
    
    \begin{subfigure}[b]{0.8\textwidth}
        \centering
        \includegraphics[width=\textwidth]{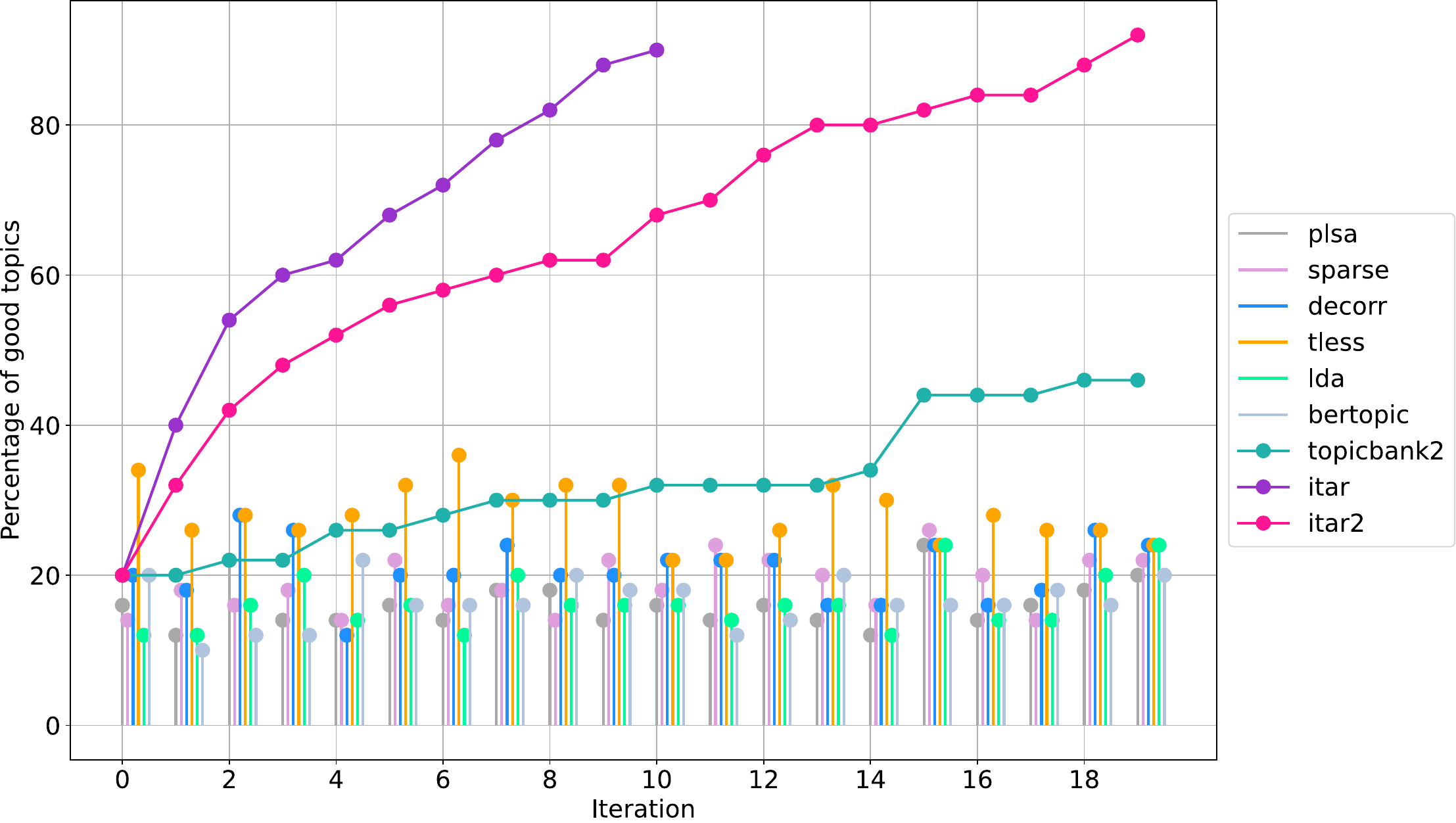}
        \caption{%
            20Newsgroups, $T \hm= 50$.
        }
        \label{fig:toptok-plots-20ng}
    \end{subfigure}

    \caption{%
        Percentage of good model topics depending on iteration ($\uparrow$). In iterative models (TopicBank2, ITAR, ITAR2), each subsequent model is trained based on the previous one, hence the monotonic dependence (in contrast to non-iterative models).
    }
    \label{fig:toptok-plots}
\end{figure}

\begin{table}[!ht]
    \centering
    \caption{%
        Some properties of the final models: perplexity ($\PPL$, $\downarrow$), average topic coherence ($\Coh \hm\equiv \Cohtok$, $\uparrow$), percentage of good topics ($\Tplus$, $\uparrow$), topic diversity ($\Div$, $\uparrow$). Results for two experiments are presented: PostNauka, models for $T \hm= 20$ topics (left); RuWiki-Good, models for $T \hm= 50$ topics (right). It can be seen that the ITAR and ITAR2 iterative models have the highest percentage of good topics ($\mathrm T_{+}$). At the same time, the topics are diverse ($\Div$) and the perplexity of the whole model is moderate ($\PPL$).
        For BERTopic, TopicBank, and TopicBank2, the column for perplexity contains two values in the format $\PPL_1/\PPL_2$: the second one $\PPL_2$ is ``honest'' perplexity, calculated as for a topic model with exactly the same topics as in BERTopic, TopicBank and TopicBank2, respectively. The first perplexity value $\PPL_1$, on the other hand, is calculated by adding one additional background topic to the topics of the aforementioned models (the topic which is known to be bad, but which is not taken into account when calculating other model quality indicators). This gives the BERTopic, TopicBank, and TopicBank2 models more freedom and, in our opinion, makes the perplexity comparison with other topic models more ``interesting''.
    }  
    \label{tab:toptok-results}
    \begin{tabular}{l|cccc|cccc}
        \toprule
        \multirow{2}{*}{\textbf{Model}} & \multicolumn{4}{c|}{\textbf{PostNauka (20 topics)}} & \multicolumn{4}{c}{\textbf{RuWiki-Good (50 topics)}} \\
        \cline{2-9}
                      & $\PPL$/1000  & $\Coh$  & $\Tplus$, \%  & $\Div$  & $\PPL$/1000  & $\Coh$  & $\Tplus$, \%  & $\Div$  \\
        \midrule
        plsa          & \textbf{2.99} & 0.74          & 20          & 0.60          & \textbf{3.46}  & 0.81          & 26          & 0.66 \\
        sparse        & 3.33          & 0.84          & 40          & 0.66          & 3.85           & 0.85          & 28          & 0.68 \\
        decorr & 3.15          & 0.79          & 40          & 0.61          & 3.62           & 0.86          & 30          & 0.67 \\  
        tless         & 3.65          & 0.75          & 30          & \textbf{0.75} & 4.98           & 0.71          & 24          & 0.72 \\
        lda           & \textbf{2.99} & 0.73          & 25          & 0.58          & \textbf{3.48}  & 0.83          & 24          & 0.65 \\
        bertopic      & 4.26/5.93     & \textbf{1.16} & 75          & 0.67          & 3.17/5.06      & \textbf{1.34} & 70          & 0.67 \\
        topicbank     & 4.22/6.11     & 0.98          & 30          & 0.60          & 7.39/12.94     & \textbf{1.33} & 20          & 0.68 \\
        topicbank2    & 4.12/8.11     & \textbf{1.10} & 70          & 0.67          & 6.09/11.30     & 1.16          & 44          & 0.69 \\
        itar  & 3.79          & 1.02          & \textbf{90} & \textbf{0.76} & 4.62           & 1.12          & \textbf{86} & \textbf{0.77} \\
        itar2 & 3.75          & 1.00          & \textbf{90} & 0.74          & 4.53           & 1.23          & \textbf{96} & \textbf{0.77} \\
        \bottomrule
    \end{tabular}
\end{table}

\subsection{Ablation Study}

Iterative model updating is provided by applying several regularizers: the one fixing good topics~\eqref{eq:r-fix}, the regularizer filtering collected bad topics~\eqref{eq:r-decorr1}, and the one filtering collected good topics.
But what is the contribution of each of the regularizers to the final quality of the iterative topic model?
Are all the regularizers listed equally important?

To find out this, a separate experiment was conducted, which consisted in training several iterative models besides the ``full-fledged'' ITAR, in each of which one or two regularizers were ``disabled''.
The results obtained for the PostNauka dataset at $T \hm= 20$ are summarized in Tab.~\ref{tab:toptok-ablation-results}.
From where the contribution of each of the regularizers can be seen:
fixing good topics expectedly increases the final percentage of good topics (as well as the perplexity); filtering out bad ones reduces the frequency of bad topics in models trained on separate iterations; filtering out good topics leads to more diverse topics.

\begin{table}[!ht]
    \centering
    \caption{%
        The effect of different parts of the ITAR model on the final result using the PostNauka dataset as an example when training models on $T \hm= 20$ topics. The name format is ``itar\_[is there fixation of good topics]-[is there filtering out of bad topics]-[is there filtering out of good topics]''. The ``\# iters''~($\downarrow$) column shows how many iterations the training took (as a percentage of the maximum number of $20$ iterations). The $\Tminus$~($\downarrow$) column, as a percentage of the number of topics in one model ($T \hm= 20$), shows the total number of bad topics found by the iterative model over all training iterations, from the first to the last (as opposed to the $\Tplus$ column, which shows the percentage of good topics in the model at the last final iteration only). The columns $\PPL$, $\Coh$, $\Tplus$, $\Div$ have the same meaning as in Tab.~\ref{tab:toptok-results}. $\Tplus$ shows that fixing good topics expectedly increases the proportion of good topics in the model; $\Tminus$ shows that filtering out bad topics reduces the frequency of bad topics appearance; and $\Div$ shows that filtering out good topics also leads to more different topics.
    }
    \label{tab:toptok-ablation-results}
    \begin{tabular}{l|cccccc}
        \toprule
        \multirow{2}{*}{\textbf{Model}} & \multicolumn{6}{c}{\textbf{PostNauka (20 topics)}} \\ \cline{2-7}
            & \# iters, \%  & $\PPL$/1000  & $\Coh$  & $\Tplus$, \%  & $\Tminus$, \%  & $\Div$  \\  
        \midrule
        itar        & \textbf{50} & 3.79          & \textbf{1.02} & \textbf{90} & \textbf{100} & \textbf{0.76} \\
        itar\_0-0-1 & 85          & \textbf{3.30} & 0.81          & 35          & 275          & 0.66 \\
        itar\_0-1-0 & \textbf{60} & \textbf{3.31} & 0.86          & 50          & 350          & 0.71 \\
        itar\_0-1-1 & 85          & \textbf{3.31} & 0.93          & 50          & 325          & 0.71 \\
        itar\_1-0-0 & 70          & 3.56          & 0.90          & 60          & 230          & 0.69 \\
        itar\_1-0-1 & 90          & 3.65          & 0.95          & 75          & 200          & 0.72 \\
        itar\_1-1-0 & 90          & 3.75          & \textbf{1.05} & \textbf{95} & \textbf{95}  & \textbf{0.75} \\
        \bottomrule
    \end{tabular}
\end{table}

\subsection{Intra-Text Coherence}

The calculation of intra-text coherence is supposed to be performed on text with natural word order~\cite{intracoh}, so among all datasets (see Tab.~\ref{tab:datasets}) we used only those where lemmatized text in natural word order was available.

When evaluating topic goodness by its intra-text coherence value, the iterative model also improves monotonically, but ends up being comparable to the best non-iterative models (see Fig.~\ref{fig:intratext-plot}).
The point is that the value of a topic's intra-text coherence, unlike its top-word coherence, \emph{depends on other topics}.
Thus, fixing good topics by intra-text coherence (those that occur in the text in long homogeneous segments) restricts the freedom of the model even more strongly than fixing topics selected by word co-occurrences (there just can not be too many topics with a large number of long segments).

%
%
%
It has been observed that increasing regularization (when more and more topics are fixed) can lead to intra-text coherence for some topics becoming zero (meaning that for no word in the text, the maximum among probabilities $p(t \hm\mid w)$ falls on those topics~\cite{intracoh}).
Therefore, in addition to the stopping criterion based on the number of good topics, as in the experiment where topic quality was assessed by Newman coherence~(see Sec.~\ref{sec:toptok-results}), the following criterion was applied: training of the iterative model was stopped if at least one topic had zero intra-text coherence.

\begin{figure}[!ht]
    \centering
    \begin{subfigure}[b]{0.8\textwidth}
        \centering
        \includegraphics[width=\textwidth]{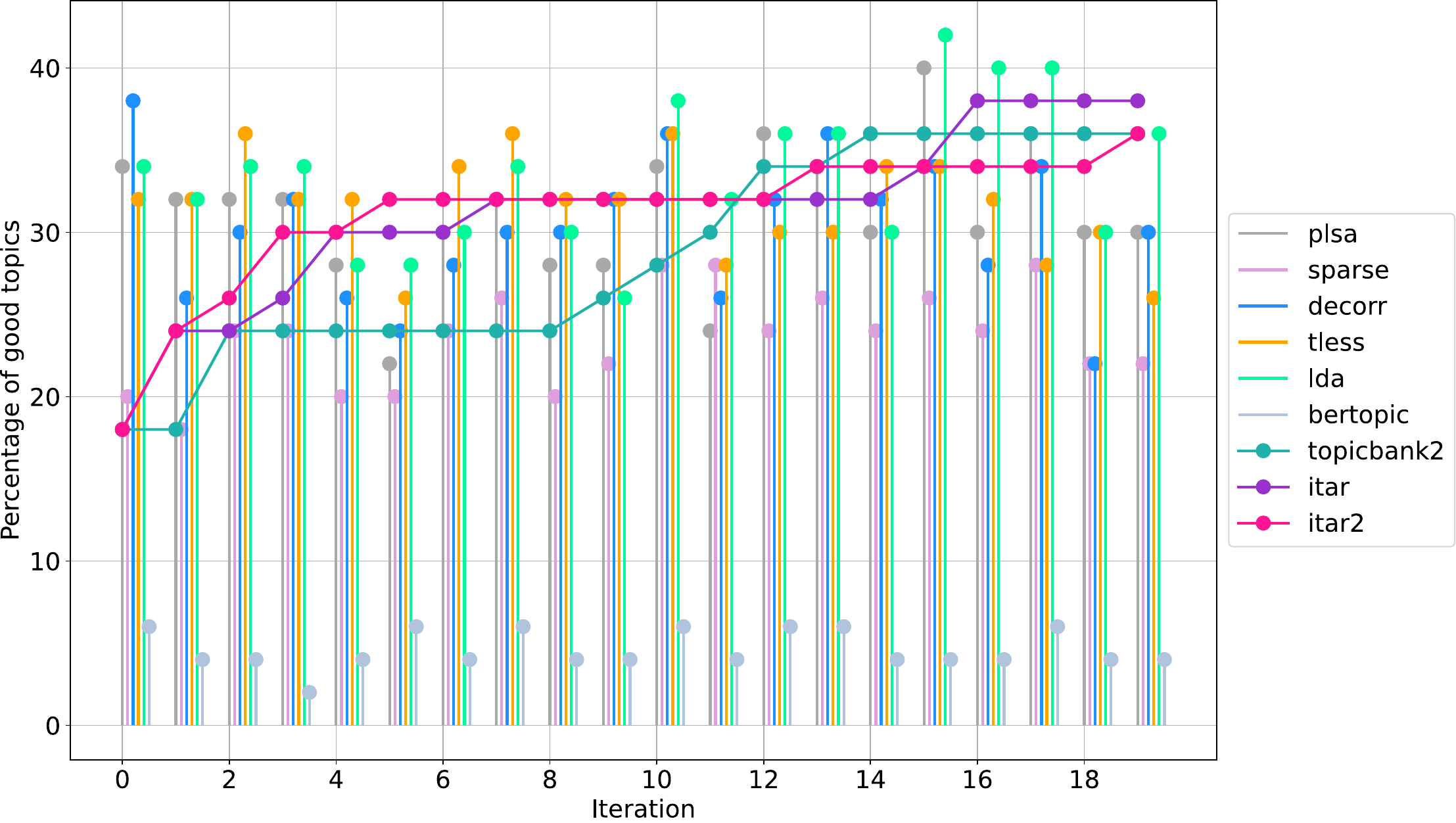}
        \caption{%
            20Newsgroups, $T \hm= 50$.
        }
    \end{subfigure}

    \begin{subfigure}[b]{0.8\textwidth}
        \centering
        \includegraphics[width=\textwidth]{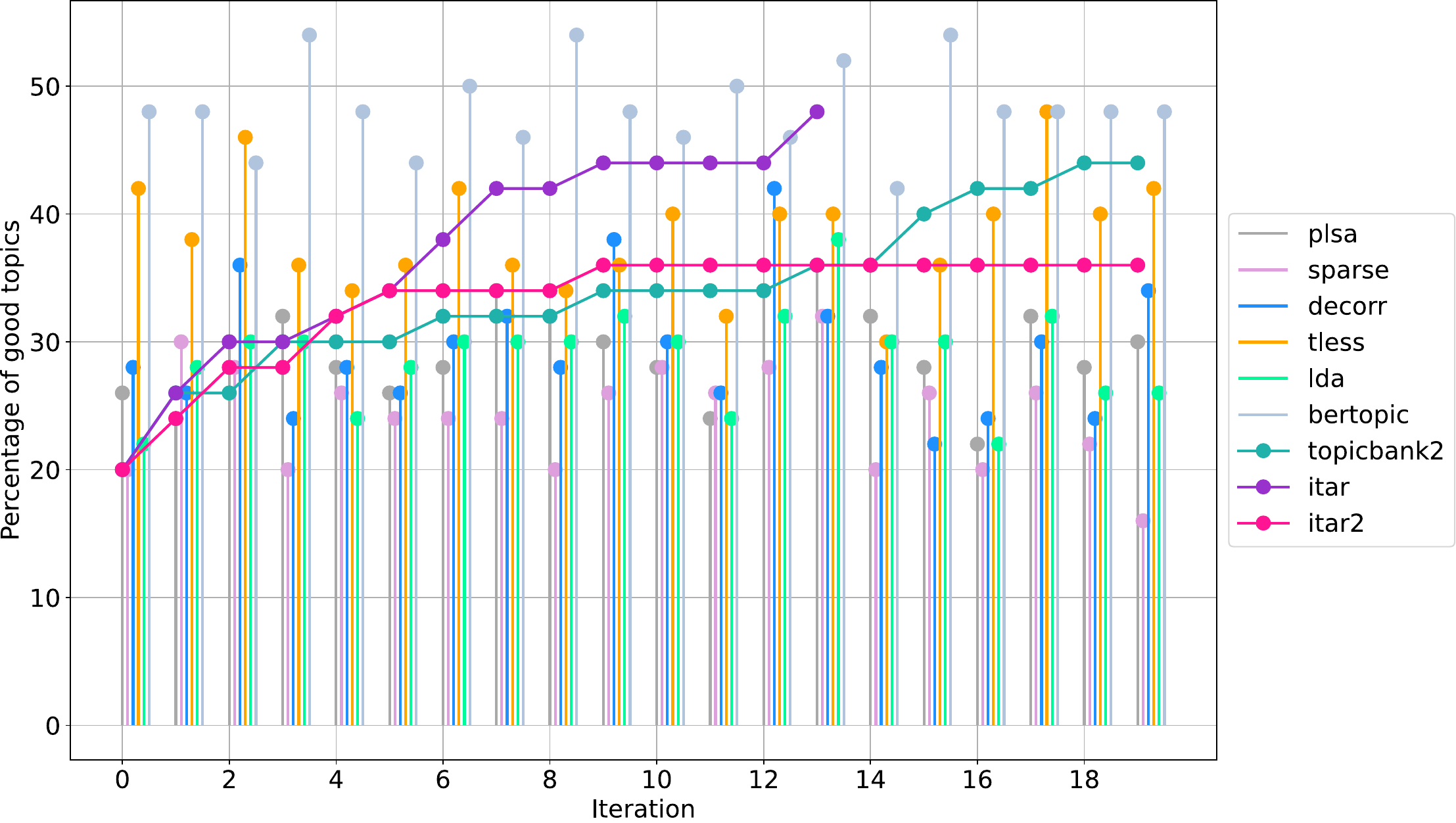}
        \caption{%
            ICD-10, $T \hm= 50$.
        }
    \end{subfigure}

    \caption{%
        Percentage of good topics in the model as a function of iteration ($\uparrow$). In contrast to the results shown in Fig.~\ref{fig:toptok-plots}, the goodness of a topic was determined by the value of its intra-text coherence, rather than by the coherence of top-word co-occurrences. Since the intra-text coherence scores of different topics are not independent, in this case, it is more difficult for the iterative model to accumulate good topics.
        (As can be seen in the graph for ITAR2 model, when more than half of the iterations passed without adding new topics at all.
        Moreover, it can be seen that the TopicBank2 model could perform better than ITAR2, because in TopicBank the models are trained independently of each other at different iterations, and therefore collected good topics do not influence the quality assessment of new topics; in ITAR2, pairwise correlation with collected good topics is also applied, which further narrows the search area for new topics.
        The graph for ITAR stops before reaching maximum iteration because so many good topics were accumulated that their fixation by regularization led to the degeneration of the remaining free topics.)
    }
    \label{fig:intratext-plot}
\end{figure}

In experiments with different numbers of topics in the models ($20$ and $50$), the relative coherence thresholds by which good and bad topics were determined remained the same~(see Sec.~\ref{sec:methodology}).
However, the absolute thresholds appeared different.
In the case of Newman coherence for all datasets~(\ref{sec:data}), there was an increase in absolute thresholds when the number of topics in the models changed from $20$ to $50$, while in the case of intra-text coherence, there was a decrease.
The latter is explained by the fact that topics in the models are assumed to be equivalent~\cite{veselova2020topic}, and increasing the number of topics in the model leads to a decrease in the size of topics,\footnote{Topic size, or topic capacity, is the amount of text occupied by a topic~\cite{veselova2020topic}: $n_t \hm\equiv \sum_{d \in D} n_d \theta_{td}$. There is also a slightly different way to define the topic size which follows directly from the notations introduced in section~\ref{sec:artm}: $n_t \hm= \sum_{d \in D} n_{td} = \sum_{d \in D} \sum_{w \in d} n_{dw} p_{tdw}$.} hence a decrease in the average length of the topic segment.
The increase in Newman coherence, on the other hand, can be explained as follows.
Thus, the decrease in the size of topics is partly due to the splitting of larger topics into smaller ones~\cite{alekseev2021topicbank}.
And Newman's coherence will increase if one large heterogeneous topic (which does not have very high top-word co-occurrences) splits into several smaller, but already more homogeneous topics (the ones that have the top words more compactly distributed in the text).

Iterative models trained to accumulate intra-text coherent topics also appear to contain a high percentage of top-word coherent topics (see Tab.~\ref{tab:cohs-relation}).
Moreover, such topics are not just abundant in the model as a whole, but a high percentage of Newman-coherent topics are also contained among the intra-text coherent topics themselves (see Tab.~\ref{tab:cohs-relation2}).
This indicates the positive relationship between the two measures of topic coherence---different approaches to estimating topic interpretability.

\begin{table}[!ht]
    \centering
    \caption{%
        The relationship between top-word co-occurrence coherence and intra-text coherence estimated by the topics of individual models. For iterative models (TopicBank2, ITAR, ITAR2), the following metrics are presented for the first and last training iterations: $\Tplustoptok$~($\uparrow$) means \emph{relative density of topics that are good in terms of top-word co-occurrence coherence $\Cohtok$ among the topics in the model}. For example, if there are $2$ good topics per $20$ topics in the model, then the density of good topics is $2 \hm/ 20 \hm= 10\%$. If we know that the average density of good topics in trained models is $20\%$, then the relative density of good topics for the example model is $10 \hm/ 20 \hm= 0.5$. Thus, if the relative density of good topics is higher than one, it means that this model contains more good topics on average than all the topic models studied. Further, $\Tplustoptokatintra$~($\uparrow$) refers to the relative density of good topics concerning co-occurrence coherence among model topics that are good in terms of intra-text coherence. For example, if a $20$-topic model has $2$ topics that are good by intra-text coherence, of which only $1$ is also good by Newman PMI-based coherence, then the value of $\Tplustoptokatintra$ for this model is $(1 \hm/ 2) \hm/ 0.2 \hm= 2.5$. Similarly, the higher this density is above one, the more among the intra-text coherent topics that are also coherent in terms of top-word co-occurrences.
        Based on the experiment setting (see Sec.~\ref{sec:methodology}), the average percent of good topics in the model, as judged by $\Cohtok$, is $20\%$, and therefore the average $\Tplustoptok$ is equal to $1.0$.
        From the experiment results, the average $\Tplustoptok$ value for the model topics is equal to $1.0 \hm\pm 0.3$ (the deviation depends on the dataset and the number of topics in the models, so it is unique for each experiment; and from what we observed, $0.3$ can on average serve as a good estimate of the deviation).
        TopicBank2 has low initial $\Tplustoptok$ mainly just because after the first iteration there is usually yet a small number of topics in the topic bank (TopicBank as a topic model gradually increases the number of topics).
    }
    \label{tab:cohs-relation}
    \begin{tabular}{l|cc|cc}
        \toprule
        \multirow{2}{*}{\textbf{Model}} & \multicolumn{2}{c}{First iteration} & \multicolumn{2}{c}{Last iteration} \\
            & $\Tplustoptok$ & $\Tplustoptokatintra$ & $\Tplustoptok$ & $\Tplustoptokatintra$ \\
        
        \midrule
        \multicolumn{5}{c}{\textbf{PostNauka (20 topics)}}\\
        \midrule
        
        topicbank2 & 0.5 & \multirow{3}{*}{\textbf{3.3}} & 0.5           & \textbf{1.7} \\
        itar       & \multirow{2}{*}{1.5} &                                & \textbf{2.8} & \textbf{2.1} \\
        itar2      &                      &                                & \textbf{2.5}  & \textbf{2.1} \\
        
        \midrule
        \multicolumn{5}{c}{\textbf{20Newsgroups (50 topics)}}\\
        \midrule
        
        topicbank2 & 0.4 & \multirow{3}{*}{\textbf{2.2}} & 0.6          & \textbf{1.7} \\
        itar       & \multirow{2}{*}{1}   &                                & \textbf{1.7} & \textbf{2.1} \\
        itar2      &                      &                                & \textbf{1.4} & \textbf{1.7} \\
        
        \bottomrule
    \end{tabular}
\end{table}




\begin{table}[!ht]
    \centering
    \caption{%
        The relationship between top-word co-occurrence coherence and intra-text coherence estimated by the total set of topics.
        For datasets with natural word order and explored numbers of topics $T$ in one model, $\Tplustoptokatintra$ refers to the relative density of top-word coherent topics among intra-text coherent ones.
        However, in this table, the density values are not \emph{local}, for topics of individual models (like in Tab.~\ref{tab:cohs-relation}), but \emph{global}: estimated for all accumulated topics of all models, which were used in the calculation of absolute coherence thresholds to determine whether a model topic is good or bad (see Sec.~\ref{sec:methodology}).
        The expected value of the global density $\Tplustoptok$ of top-word coherent topics among all topics is also $1.0$.
        The experimentally calculated values of $\Tplustoptok$ are $1.00 \hm\pm 0.07$ for $T \hm= 20$, and $1.00 \hm\pm 0.05$ for $T \hm= 50$.
        (The deviations were estimated by repeatedly sampling from the total set of topics obtained from all models, a few in the number $\mathrm T_+^{intra}$ and calculating the density $\Tplustoptok$ on this subsample, where $\mathrm T_+^{intra}$ is the total number of topics coherent over $\Cohintra$.)
        Thus, values of $\Tplustoptokatintra$ greater than one indicate that among the intra-text coherent topics, there is, on average, a large number of Newman-coherent topics.
    }
    \label{tab:cohs-relation2}

    \begin{tabular}{l|cc}
        \toprule
        \multirow{2}{*}{\textbf{Dataset}} & 20 topics & 50 topics \\
            & \multicolumn{2}{c}{$\Tplustoptokatintra$}\\
        \midrule
        PostNauka & 2.11 & 1.73 \\
        20Newsgroups & 1.18 & 1.71 \\
        ICD-10 & 1.53 & 1.85 \\
        \bottomrule
    \end{tabular}
\end{table}

\section{Conclusion}

The paper presents an iteratively updated topic model as a series of related topic models trained one after another.
The process is designed so that the iterative model accumulates already found and seeks new good topics.
Iterative update of the model is implemented within the ARTM framework: a topic fixing regularizer (smoothing-like) is responsible for preserving good topics, and a filtering regularizer (decorrelation-like) of collected good and bad topics contributes to finding new good ones.
Thus, the connection between models is done by regularization, hence the name: iterative additively regularized topic model (ITAR).

Experiments have been conducted on several collections of natural language texts to compare ITAR with other topic models.
It is shown that the iterative model presented outperforms all other ones in terms of the number of good topics, where goodness is determined by coherence based on top-word co-occurrences (PMI); its topics are diverse and its perplexity is moderate.

\section{Limitations}



It is worth noting a few limitations of the research process and/or inherent in the result obtained.

The application of an iterative approach to improving the topic model implies the use of the ARTM framework.
Thus, the question of whether it is possible to iteratively improve the BERTopic model, for example, has not been investigated.
Still, we can note that BERTopic does have some opportunities for (semi-)supervised topic modelling.
The one which best resembles the proposed fixing regularizer for the ITAR model is ``guided topic modeling''.
It is a technique which allows to ``guide'' a BERTopic model in a desired direction (hence the name) by providing a set of so-called ``seed topics'' (just as word sequences, not as probability distributions over words).
However, these seed topics are not (strictly) preserved.\footnote{{\UrlFont\url{https://maartengr.github.io/BERTopic/getting_started/guided/guided.html}}.}

The accumulation of topics is not possible for all criteria.
The iterative model is effective if the criterion of one topic's goodness does not depend on other topics.


Obtaining an iterative model requires multiple-model training.
If the dataset is big, this may not be efficient.
(Even if the dataset is not big, it is still cumbersome.)

Only automatically computed coherence criteria have been used as a measure of the goodness of topics.
However, human evaluation is more reliable.
(But at the same time more expensive and harder to obtain.)

No visible difference was found between ITAR and ITAR2 models.
In all cases considered, both showed similar results (thus the simpler ITAR model can be recommended).
However, it seems that when the number of collected bad topics is large, the regularizer~\eqref{eq:r-decorr1} involved in the ITAR model will not be able to filter out bad topics effectively.

Based on the above, it is possible to note some possible directions for further research on the development of the approach:
\begin{itemize}
    \item Compare more thoroughly and systematically the effectiveness of filtering out bad topics by regularizers~\eqref{eq:r-decorr1} and \eqref{eq:r-decorr2}.
    \item Speed up iterative model training (it would be ideal to train a good model at once).
    \item Increase the average coherence of the topics collected in the iterative model.
    \item Explore the possibility of using other criteria for topic selection (e.g., human or LLM evaluation~\cite{yang2024llm}).
    \item Explore the possibility of reducing the perplexity of a trained iterative model.
    \item Investigate whether it is possible to get all 100\% of good topics in a model.
    \item Creating intra-text coherence that is calculated for a given topic independently of other topics (or at least not as dependently as in~\cite{intracoh}).
\end{itemize}

\begin{credits}

\subsubsection{\ackname}

Thanks to Taya for the productive reflections and the opportunity to stop and just look at the world around.

\subsubsection{\discintname}
The authors have no competing interests to declare that are
relevant to the content of this article.

\end{credits}

\bibliographystyle{splncs04}
\bibliography{biblio}

\end{document}